\documentclass[runningheads]{llncs}
\usepackage{graphicx}
\usepackage{amsmath}
\usepackage{booktabs}
\usepackage{graphicx}
\usepackage{cite}
\usepackage{times}
\usepackage{epsfig}
\usepackage{dsfont}
\usepackage{amsmath}
\usepackage{amssymb}
\usepackage{multirow}
\usepackage[normalem]{ulem}
\usepackage{url}
\usepackage{xcolor}
\usepackage{dirtytalk}
\usepackage[subrefformat=parens,labelformat=parens]{subcaption}

\begin{document}
\title{Constrained Contrastive Distribution Learning for Unsupervised Anomaly Detection and Localisation in Medical Images}

%
%
\author{Yu Tian\inst{1,3}$\quad$
Guansong Pang \inst{1} $\quad$
Fengbei Liu \inst{1} $\quad$
Yuanhong chen\inst{1}$\quad$ \\
Seon Ho Shin\inst{2}$\quad$ 
Johan W. Verjans\inst{1,2,3}$\quad$
Rajvinder Singh\inst{2}$\quad$
Gustavo Carneiro\inst{1}}
%


\institute{Australian Institute for Machine Learning, University of Adelaide \and
Faculty of Health and Medical Sciences, University of Adelaide  \and
South Australian Health and Medical Research Institute
}
\maketitle              
\begin{abstract}


Unsupervised anomaly detection (UAD) learns one-class classifiers exclusively with normal (i.e., healthy) images to detect any abnormal (i.e., unhealthy) samples that do not conform to the expected normal patterns. UAD has two main advantages over its fully supervised counterpart. Firstly, it is able to directly leverage large datasets available from health screening programs that contain mostly normal image samples, avoiding the costly manual labelling of abnormal samples and the subsequent issues involved in training with extremely class-imbalanced data. Further, UAD approaches can potentially detect and localise any type of lesions that deviate from the normal patterns. One significant challenge faced by UAD methods is how to learn effective low-dimensional image representations to detect and localise subtle abnormalities, generally consisting of small lesions. To address this challenge, we propose a novel self-supervised representation learning method, called \underline{C}onstrained \underline{C}ontrastive \underline{D}istribution learning for anomaly detection (CCD), which learns fine-grained feature representations by simultaneously predicting the distribution of augmented data and image contexts using contrastive learning with pretext constraints. The learned representations can be leveraged to train more anomaly-sensitive detection models. Extensive experiment results show that our method outperforms current state-of-the-art UAD approaches on three different colonoscopy and fundus screening datasets. Our code is available at \url{https://github.com/tianyu0207/CCD}.

\end{abstract}
\section{Introduction}
\label{sec:introduction}

Classifying and localising malignant tissues have been vastly investigated in medical imaging~\cite{tian2019one,tian2020few,liuphotoshopping,litjens2017survey,baur2020scale,fan2020pranet,lz2020computer,liu2021self,liu2021noisy}.
Such systems are useful in health screening programs that require radiologists to analyse large quantities of images~\cite{tian2021detecting,pu2019prospective}, where the majority contain normal (or healthy) cases, and a small minority have abnormal (or unhealthy) cases that can be regarded as anomalies.
Hence, to avoid the difficulty of learning from such class-imbalanced training sets and the prohibitive cost of collecting large sets of manually labelled abnormal cases, several papers investigate anomaly detection (AD) with a few or no labels as an alternative to traditional fully supervised imbalanced learning~\cite{liuphotoshopping,baur2020scale,F-anoGAN,seebock2019exploiting,tian2021weakly,luo2020encoding,tian2020few,pang2021deep,uzunova2019unsupervised,ouardini2019towards}.
UAD methods typically train a one-class classifier using data from the normal class only, and anomalies (or abnormal cases) are detected based on the extent the images deviate from the normal class.


Current anomaly detection approaches~\cite{F-anoGAN,gong2019memorizing,chen2021unsupervised,liu2019photoshopping,tian2020few,venkataramanan2020attention,chen2020unsupervised} train deep generative models (e.g., auto-encoder~\cite{kingma2013auto}, GAN~\cite{goodfellow2014generative}) to reconstruct normal images, 
and anomalies are detected from the reconstruction error~\cite{pang2021deep}.
These approaches rely on a low-dimensional image representation that must be effective at reconstructing normal images, where the main challenge is to detect anomalies that show subtle deviations from normal images, such as with small lesions~\cite{tian2020few}. Recently, self-supervised methods that learn auxiliary pretext tasks~\cite{hendrycks2019using,golan2018deep,bergman2020classification,simclr,moco,liu2020self}
have been shown to learn effective representations for UAD in general computer vision tasks~\cite{hendrycks2019using,golan2018deep,bergman2020classification}, so it is important to investigate if self-supervision can also improve UAD for medical images. 

The main challenge for the design of UAD methods for medical imaging resides in how to devise effective pretext tasks.
Self-supervised pretext tasks consist of predicting geometric or brightness transformations~\cite{hendrycks2019using,golan2018deep,bergman2020classification}, or contrastive learning~\cite{simclr,moco}.
These pretext tasks have been designed to work for downstream classification problems that are not related to anomaly detection, so
they may degrade the detection performance of UAD methods~\cite{wang2020understanding}.
Sohn et al.~\cite{sohn2020learning} tackle this issue
by using smaller batch sizes than in~\cite{simclr,moco} and a new data augmentation method. 
However, the use of self-supervised learning in UAD for medical images has not been investigated, to the best of our knowledge. Further, although transformation prediction and contrastive learning show great success in self-supervised feature learning, there are no studies on how to properly combine these two approaches to learn more effective features for UAD.

In this paper, we propose \underline{C}onstrained \underline{C}ontrastive \underline{D}istribution learning (CCD), a new self-supervised representation learning designed specifically to learn normality information from exclusively normal training images. The contributions of CCD are: a) contrastive distribution learning, and b)two pretext learning constraints, both of which are customised for anomaly detection (AD). Unlike modern self-supervised learning (SSL)~\cite{simclr,moco} that focuses on learning generic semantic representations for enabling diverse downstream tasks, CCD instead contrasts the distributions of strongly augmented images (e.g., random permutations). The strongly augmented images resemble some types of abnormal images, so CCD is enforced to learn discriminative normality representations by its contrastive distribution learning. The two pretext learning constraints on augmentation and location prediction are added to learn fine-grained normality representations for the detection of subtle abnormalities. These two unique components result in significantly improved self-supervised AD-oriented representation learning, substantially outperforming previous general-purpose SOTA SSL approaches~\cite{simclr,hendrycks2019using,golan2018deep,bergman2020classification}. Another important contribution of CCD is that it is agnostic to downstream anomaly classifiers. We empirically show that our CCD improves the performance of three diverse anomaly detectors (f-anogan~\cite{F-anoGAN}, IGD~\cite{chen2021unsupervised}, MS-SSIM)~\cite{wang2003multiscale}). Inspired by IGD~\cite{chen2021unsupervised}, we adapt our proposed CCD pretraining on global images and local patches, respectively.  Extensive experimental results on three different health screening medical imaging benchmarks, namely, colonoscopy images from two datasets~\cite{borgli2020hyperkvasir,liu2019photoshopping}, and fundus images for glaucoma detection~\cite{li2019attention}, show that our proposed self-supervised approach enables the production of SOTA anomaly detection and localisation in medical images.

\section{Method}

\begin{figure}
\begin{center}
\includegraphics[width=0.88 \textwidth]{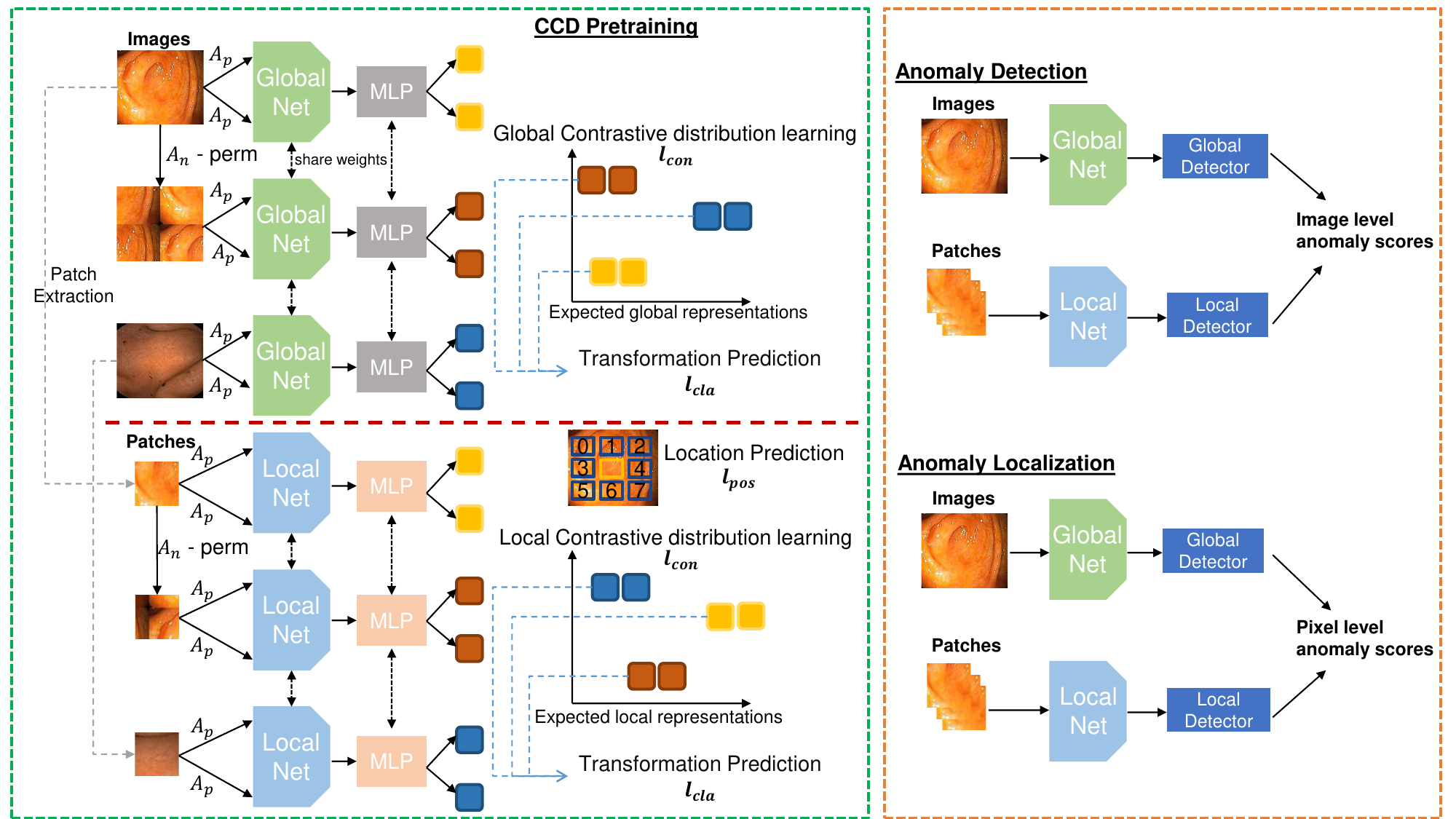}
\end{center}
   \caption{Our proposed CCD framework. \textbf{Left} shows the proposed pre-training method that unifies a contrastive distribution learning and pretext learning on both global and local perspectives (Sec.~\ref{sec:contrastive_and_pretext}),  \textbf{Right} shows the inference for detection and localisation (Sec.~\ref{sec:anomaly_detection_localisation}).}
   
\label{fig:framework}
\end{figure}

In this section, we introduce the proposed approach, depicted in the diagram of Fig,~\ref{fig:framework}. 
Specifically, given a training medical image dataset $\mathcal{D} = \{ \mathbf{x}_i \}_{i=1}^{|\mathcal{D}|}$, 
with all images assumed to be from the $\text{normal}$ class and $\mathbf{x} \in \mathcal{X} \subset \mathbb{R}^{H \times W \times C}$, our approach aims to learn anomaly detection and localisation using three modules: 1) a self-supervised constrained contrastive feature learner that pre-trains an encoding network $f_{\theta}:\mathcal{X} \rightarrow \mathcal{Z}$ (with $\mathcal{Z} \subset \mathbb{R}^{d_{z}}$) tailored for anomaly detection, 2) an anomaly classification model $h_{\psi}:\mathcal{Z} \rightarrow [0,1]$ that is built upon the pre-trained network, and 3) an anomaly localiser that leverages the classifier $h_{\psi}(f_{\theta}(\mathbf{x}_{\omega}))$ to localise an abnormal image region $\mathbf{x}_{\omega} \in \mathbb{R}^{\hat{H} \times \hat{W} \times C}$, centred at $\omega \in \Omega$ ($\Omega$ is the image lattice) with height $\hat{H} << H$ and width $\hat{W} << W$. 
The approach is evaluated on a testing set $\mathcal{T} = \{ (\mathbf{x},y,\mathbf{m})_i \}_{i=1}^{|\mathcal{T}|}$, where $y \in \mathcal{Y} = \{\text{normal}, \text{abnormal} \}$, and $\mathbf{m}\in \mathcal{M} \subset \{0,1\}^{H \times W \times C}$ denotes the segmentation mask of the lesion in the image $\mathbf{x}$. For adapting our CCD pretraining on patch representations, we simply crop the training images into patches before applying our method. 



\subsection{Constrained Contrastive Distribution Learning}
\label{sec:contrastive_and_pretext}

Contrastive learning has been used by self-supervised learning methods to pre-train encoders with data augmentation~\cite{simclr,moco,wang2020understanding} and contrastive learning loss~\cite{sohn2016improved}.  The idea is to sample functions from a data augmentation distribution (e.g., geometric and brightness transformations), and assume that the same image, under separate augmentations, form one class to be distinguished against all other images in the batch~\cite{bergman2020classification,golan2018deep}.  
Another form of pre-training is based on a pretext task, such as solving jigsaw puzzle and predicting geometric and brightness transformations~\cite{simclr,moco}. 
These self-supervised learning approaches are useful to pre-train classification~\cite{simclr,moco} and segmentation models~\cite{yi2020patch,noroozi2016unsupervised}.
Only recently, self-supervised learning 
using contrastive learning~\cite{sohn2020learning} and pretext learning~\cite{golan2018deep,bergman2020classification} have been shown to be effective in anomaly detection. 
However, these two approaches are explored separately. In this paper, we aim at harnessing the power of both approaches to learn more expressive pre-trained features specifically for UAD. 
To this end, we propose the novel \underline{C}onstrained \underline{C}ontrastive \underline{D}istribution learning method (CCD).

Contrastive distribution learning is designed to enforce a non-uniform distribution of the representations in the space $\mathcal{Z}$, 
which has been associated with more effective anomaly detection performance~\cite{sohn2020learning}.
Our CCD method constrains the constrastive distribution learning with two pretext learning tasks, with the goal of enforcing further the non-uniform distribution of the representations.
The CCD loss is defined as
\begin{equation}
    \ell_{CCD}(\mathcal{D};\theta,\beta,\gamma)=
    \ell_{con}(\mathcal{D};\theta) + 
    \ell_{cla}(\mathcal{D};\beta) + 
    \ell_{pos}(\mathcal{D};\gamma), 
    \label{eq:self_supervised_training_loss}
\end{equation}
where $\ell_{con}(\cdot)$ is the contrastive distribution loss, $\ell_{cla}$ and $\ell_{pos}$ are two pretext learning tasks added to constrain the optimisation; and $\theta$, $\beta$ and $\gamma$ are trainable parameters. 
The contrastive distribution learning 
uses a dataset of \textbf{weak data augmentations} $\mathcal{A}_p=\{a_l:\mathcal{X} \to \mathcal{X}\}_{l=1}^{|\mathcal{A}_p|}$ and \textbf{strong data augmentations} $\mathcal{A}_n=\{a_l:\mathcal{X} \to \mathcal{X}\}_{l=1}^{|\mathcal{A}_n|}$, where 
$a_l(\mathbf{x})$ denotes a particular data augmentation applied to $\mathbf{x}$, and the loss is defined as
\begin{equation}
\begin{split}
    &\ell_{con}(\mathcal{D};\theta)= \\
    &-\mathbb{E} 
    \left [
    \log\frac{\exp\left[ \frac{1}{\tau} f_{\theta}(a(\tilde{\mathbf{x}}^j))^{\top} f_{\theta}(a'(\tilde{\mathbf{x}}^j)) \right]}{\exp\left[ \frac{1}{\tau} f_{\theta}(a(\tilde{\mathbf{x}}^j))^{\top} f_{\theta}(a'(\tilde{\mathbf{x}}^j)) \right] + \sum_{i=1}^M \exp\left[ \frac{1}{\tau} f_{\theta}(a(\tilde{\mathbf{x}}^j))^{\top} f_{\theta}(a'(\tilde{\mathbf{x}}_i^j)) \right]}
    \right ],
\end{split}
    \label{eq:contrastive_loss}
\end{equation}
where the expectation is over $\mathbf{x}\in\mathcal{D}$, 
$\{\mathbf{x}_i\}_{i=1}^M \subset \mathcal{D} \setminus \{\mathbf{x}\}$, $a(.),a'(.) \in  \mathcal{A}_p$, $\tilde{\mathbf{x}}^j=a_j(\mathbf{x})$,
$\tilde{\mathbf{x}}_i^j=a_j(\mathbf{x}_i)$, and $a_j(.)\in \mathcal{A}_n$. 
The images augmented with the functions from the strong set $\mathcal{A}_n$ carry some `abnormality' compared to the original images, which is helpful to learn a non-uniform distribution in the representation space $\mathcal{Z}$. 

We can then constrain further the training to learn more non-uniform representations with a self-supervised classification constraint $\ell_{cla}(\cdot)$ that enforces the model to achieve accurate classification of the strong augmentation function:
\begin{equation}
\ell_{cla}(\mathcal{D};\beta) = -\mathbb{E}_{\mathbf{x} \in \mathcal{D},  a(.) \in \mathcal{A}_n } \left [ \log \mathbf{a}^{\top} f_{\beta}(f_{\theta}(a(\mathbf{x}))) \right ],
\label{eq:pretext_loss}
\end{equation}
where $f_{\beta}:\mathcal{Z} \rightarrow [0,1]^{|\mathcal{A}_n|}$ is a fully-connected (FC) layer, and $\mathbf{a} \in \{0,1\}^{|\mathcal{A}_n|}$ is a one-hot vector representing the strong augmentation $a(.) \in \mathcal{A}_n$.

The second constraint is based on the relative patch location from the centre of the training image -- 
this positional information is important for segmentation tasks~\cite{noroozi2016unsupervised,DBLP:journals/corr/abs-1901-09005}. This constraint is added to learn fine-grained features and achieve more accurate anomaly localisation. Inspired by~\cite{doersch2015unsupervised}, the positional constraint predicts the relative position of the paired image patches, with its loss defined as 
\begin{equation}
\ell_{pos}(\mathcal{D};\gamma) = -\mathbb{E}_{ \{ \mathbf{x}_{\omega_{1}},\mathbf{x}_{\omega_{2}} \} \sim \mathbf{x} \in \mathcal{D}} \left [ \log \mathbf{p}^{\top} f_{\gamma}(f_{\theta}(\mathbf{x}_{\omega_{1}}),f_{\theta}(\mathbf{x}_{\omega_{2}})) \right ],
\label{eq:patch_prediction}
\end{equation}
where $\mathbf{x}_{\omega_{1}}$ is a randomly selected fixed-size image patch from $\mathbf{x}$, $\mathbf{x}_{\omega_{2}}$ is another image patch from one of its eight neighbouring patches (as shown in `patch location prediction' in Fig.~\ref{fig:framework}), $f_{\gamma}:\mathcal{Z}\times\mathcal{Z} \rightarrow [0,1]^8$, 
and $\mathbf{p} = \{0, 1\}^8$ is a one-hot encoding of the synthetic class label.

Overall, the constraints in~\eqref{eq:pretext_loss} and~\eqref{eq:patch_prediction} to the contrastive distribution loss in~\eqref{eq:contrastive_loss} are designed to increase the non-uniform representation distribution and to improve the representation discriminability between normal and abnormal samples, compared with~\cite{sohn2020learning}.

\subsection{Anomaly Detection and Localisation}
\label{sec:anomaly_detection_localisation}

Building upon the pre-trained encoder $f_\theta(\cdot)$ using the loss in~\eqref{eq:self_supervised_training_loss}, we fine-tune two state-of-the-art UAD methods, IGD~\cite{chen2021unsupervised} and F-anoGAN~\cite{F-anoGAN}, and a baseline method, multi-scale structural similarity index measure (MS-SSIM)-based auto-encoder~\cite{wang2003multiscale}.
All UAD methods use the same training set $\mathcal{D}$ that contains only normal image samples. 

IGD~\cite{chen2021unsupervised} combines three loss functions: 1) two reconstruction losses based on local and global multi-scale structural similarity index measure (MS-SSIM)~\cite{wang2003multiscale} and mean absolute error (MAE) to train the encoder $f_{\theta}(\cdot)$ and decoder $g_{\phi}(\cdot)$, 2) a regularisation loss to train adversarial interpolations from the encoder~\cite{berthelot2018understanding}, and 3) an anomaly  classification loss to train $h_{\psi}(\cdot)$. The anomaly detection score of image $\mathbf{x}$ is
\begin{equation}
    s_{IGD}(\mathbf{x}) = \xi \ell_{rec}(\mathbf{x},\tilde{\mathbf{x}}) + (1-\xi)(1 - h_{\psi}(f_{\theta}(\mathbf{x}))),
    \label{eq:anomaly_score_image}
\end{equation}
where $\tilde{\mathbf{x}} =g_{\phi}(f_{\theta}(\mathbf{x}))$,
$h_{\psi}(f_{\theta}(\mathbf{x})) \in [0,1]$ returns the likelihood that $\mathbf{x}$ belongs to the normal class, $\xi \in [0,1]$ is a hyper-parameter, and
\begin{equation}
    \ell_{rec}(\mathbf{x},\tilde{\mathbf{x}}) =
    \rho \| \mathbf{x} - \tilde{\mathbf{x}} \|_1 + 
     (1-\rho) \left ( 1-\left( 
     \nu m_{G}(\mathbf{x},\tilde{\mathbf{x}}) + (1-\nu)m_{L}(\mathbf{x},\tilde{\mathbf{x}}) \right ) \right ),
    \label{eq:reconstruction error}
\end{equation}
with $\rho,\nu \in [0,1]$, $m_{G}(\cdot)$ and $m_{L}(\cdot)$ denoting the global and local MS-SSIM scores~\cite{chen2021unsupervised}.
Anomaly localisation uses~\eqref{eq:anomaly_score_image} to compute $s_{IGD}(\mathbf{x}_{\omega})$, $\forall \omega \in \Omega$, where $\mathbf{x}_{\omega} \in \mathbb{R}^{\hat{H} \times \hat{W} \times C}$ is an image region--this forms a heatmap, where large values denote anomalous regions.

F-anoGAN~\cite{F-anoGAN} combines generative adversarial networks (GAN) and auto-encoder models to detect anomalies. Training involves the minimisation of reconstruction losses in both the original image and representation spaces to model $f_{\theta}(\cdot)$ and $g_{\phi}(\cdot)$. It also uses a GAN loss~\cite{goodfellow2014generative} to model $g_{\phi}(\cdot)$ and $h_{\psi}(\cdot)$.  
Anomaly detection for image $\mathbf{x}$ is 
\begin{equation}
    s_{FAN}(\mathbf{x})= \| \mathbf{x} - g_{\phi}(f_{\theta}(\mathbf{x})) \| + \kappa\| f_{\theta}(\mathbf{x})-f_{\theta}(g_{\phi}(f_{\theta}(\mathbf{x})))  \|.
\end{equation}
Anomaly localisation at $\mathbf{x}_{\omega} \in \mathbb{R}^{\hat{H} \times \hat{W} \times C}$ is achieved by $\| \mathbf{x}_{\omega} - g_{\phi}(f_{\theta}(\mathbf{x}_{\omega})) \|$, $\forall \omega \in \Omega$.

 For the MS-SSIM auto-encoder~\cite{wang2003multiscale}, we train it with the MS-SSIM loss for reconstructing the training images. 
Anomaly detection for $\mathbf{x}$ is based on  $s_{MSI}(\mathbf{x})=1-\left( 
 \nu m_{G}(\mathbf{x},\tilde{\mathbf{x}}) + (1-\nu)m_{L}(\mathbf{x},\tilde{\mathbf{x}}) \right )$, with $\tilde{\mathbf{x}}$ as defined in~\eqref{eq:anomaly_score_image}. Anomaly localisation is performed with $s_{MSI}(\mathbf{x}_{\omega})$ at image regions $\mathbf{x}_{\omega} \in \mathbb{R}^{\hat{H} \times \hat{W} \times C}$, $\forall \omega \in \Omega$. Inspired by IGD~\cite{chen2021unsupervised}, we also pretrain a local model using our CCD pretraining approach based on the local patches for F-anogan~\cite{F-anoGAN} and MS-SSIM autoencoder~\cite{wang2003multiscale}, respectively.




\section{Experiments}


\subsection{Dataset}

We test our framework on three health screening datasets. We test both anomaly detection and localisation on the colonoscopy images of Hyper-Kvasir dataset~\cite{borgli2020hyperkvasir}.
On the glaucoma datasets using fundus images~\cite{li2019attention} and colonoscopy dataset~\cite{liu2019photoshopping} that do not have lesion masks, we test anomaly detection only.
Detection is assessed with area under the ROC curve (AUC). Localisation is measured with intersection over union (ioU).

\textbf{Hyper-Kvasir} is a large multi-class public gastrointestinal dataset. The data was collected from the gastroscopy and colonoscopy procedures from Baerum Hospital in Norway. 
All labels were produced by experienced radiologists. The dataset contains 110,079 images from abnormal (i.e., unhealthy) and normal (i.e., healthy) patients, with 10,662 labelled. We use part of the clean images from the dataset to train our UAD methods. 
Specifically, 2,100 images from `cecum', `ileum' and `bbps-2-3' are selected as normal, from which we use 1,600 for training and 500 for testing. We also take 1,000 abnormal images and their segmentation masks and stored them in the testing set. 

\textbf{LAG} is a large scale fundus image dataset for glaucoma detection~\cite{li2019attention}, containing 4,854 fundus images with 1,711 positive glaucoma scans and 3,143 negative glaucoma scans. We reorganised this dataset for training the UAD methods, with 2,343 normal (negative glaucoma) images for training, and 800 normal images and 1,711 abnormal images with positive glaucoma for testing.

\textbf{Liu et al.'s colonoscopy dataset} is a colonoscopy image dataset for UAD using 18 colonocopy videos from 15
patients~\cite{liu2019photoshopping}. The training set contains 13,250 normal (healthy) images without any polyps, and the testing set contains 967 images, having 290 abnormal images with polyps and 677 normal (healthy) images without polyps. 

\subsection{Implementation Details}

For pre-training, we use Resnet18~\cite{he2016deep} as the backbone architecture for the encoder $f_{\theta}(\mathbf{x})$, and similarly to previous works~\cite{simclr,sohn2020learning}, we add an MLP to this backbone as the projection head for the contrastive learning. 
All images from the Hyper-Kvasir~\cite{borgli2020hyperkvasir} and LAG~\cite{li2019attention} datasets are resized to 256 $\times$ 256 pixels. 
For the Liu et al.'s colonoscopy dataset, images are resized to 64 $\times$ 64 pixels.  
The batch size is set to 32 and learning rate to 0.01 for the self-supervised pre-training.
We investigate the impact of different strong augmentations in $\mathcal{A}_n$ such as rotation, permutation, cutout and Gaussian noise. 
All weak augmentations in $\mathcal{A}_p$ are the same as SimCLR~\cite{simclr} (i.e., colour jittering, random grey scale, crop, resize, and Gaussian blur). 
The model is trained using SGD optimiser with  temperature 0.2. The encoder $f_{\theta}(\cdot)$ outputs a 128 dimensional feature in $\mathcal{Z}$. 
All datasets are pre-trained for 2,000 epochs. 

For the training of IGD~\cite{chen2021unsupervised}, F-anoGAN~\cite{F-anoGAN} and MS-SSIM auto-encoder~\cite{chen2021unsupervised}, we use the hyper-parameters suggested by the respective papers. 
For localisation, we compute the heatmap based on the localised anomaly scores from IGD, where the final map is obtained by summing the global and local maps. 
In our experiments, the local map is obtained by considering each 32 $\times$ 32 image patch as a instance and apply our proposed self-supervised learning to it. The global map is computed based on the whole image sized as 256 $\times$ 256. For F-anoGAN and MS-SSIM auto-encoder, we use the same setup as the IGD, where models based the 256 $\times$ 256 whole image and the 32 $\times$ 32 patches are trained, respectively.
Code will be made publicly available upon paper acceptance.

\subsection{Ablation Study}

\begin{figure}[t]
\begin{subfigure}{.5\textwidth}
\centering
  \includegraphics[width=.85\textwidth]{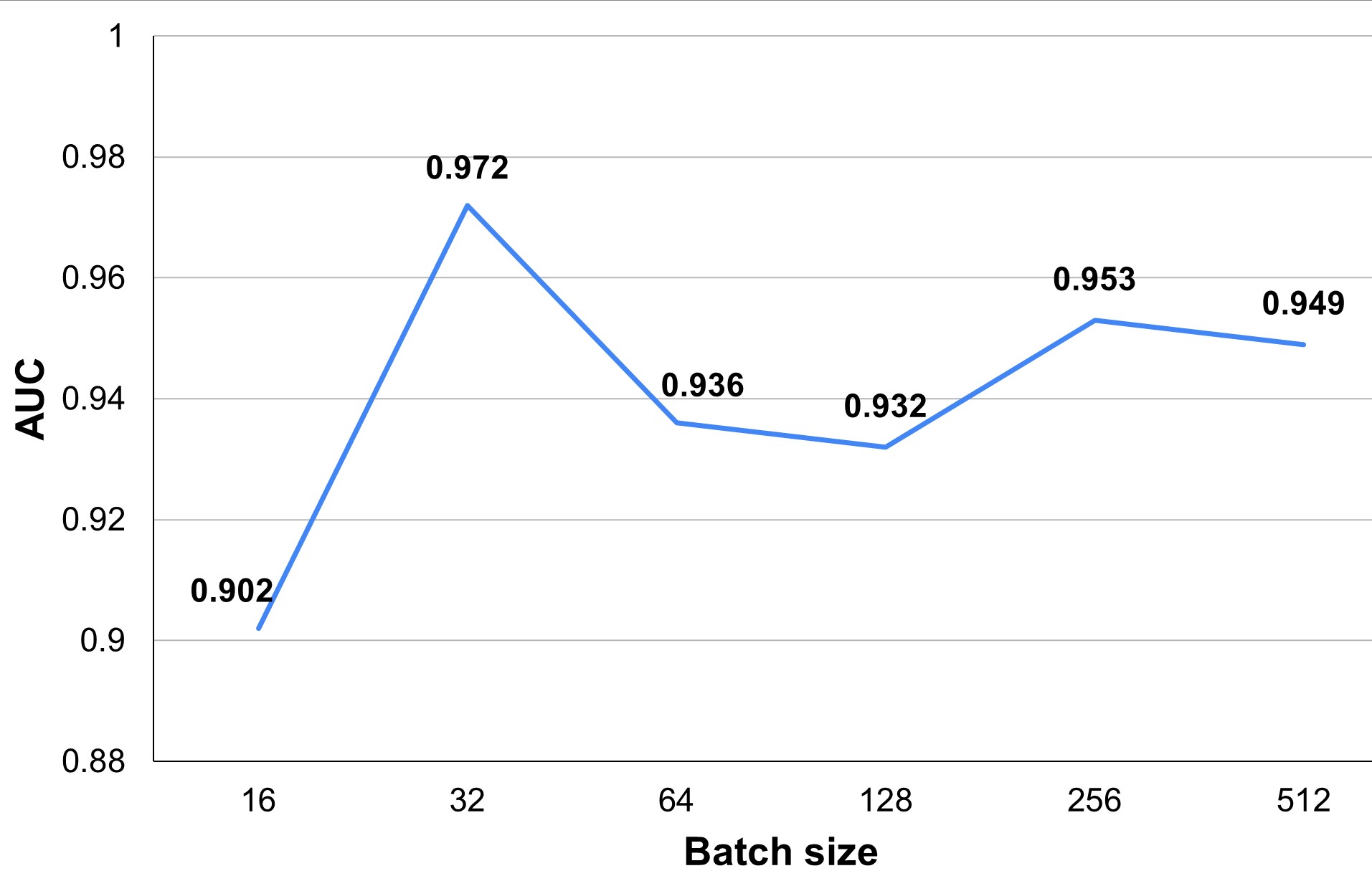}
\end{subfigure}%
\begin{subfigure}{.5\textwidth}
\centering
\includegraphics[width=0.85 \linewidth]{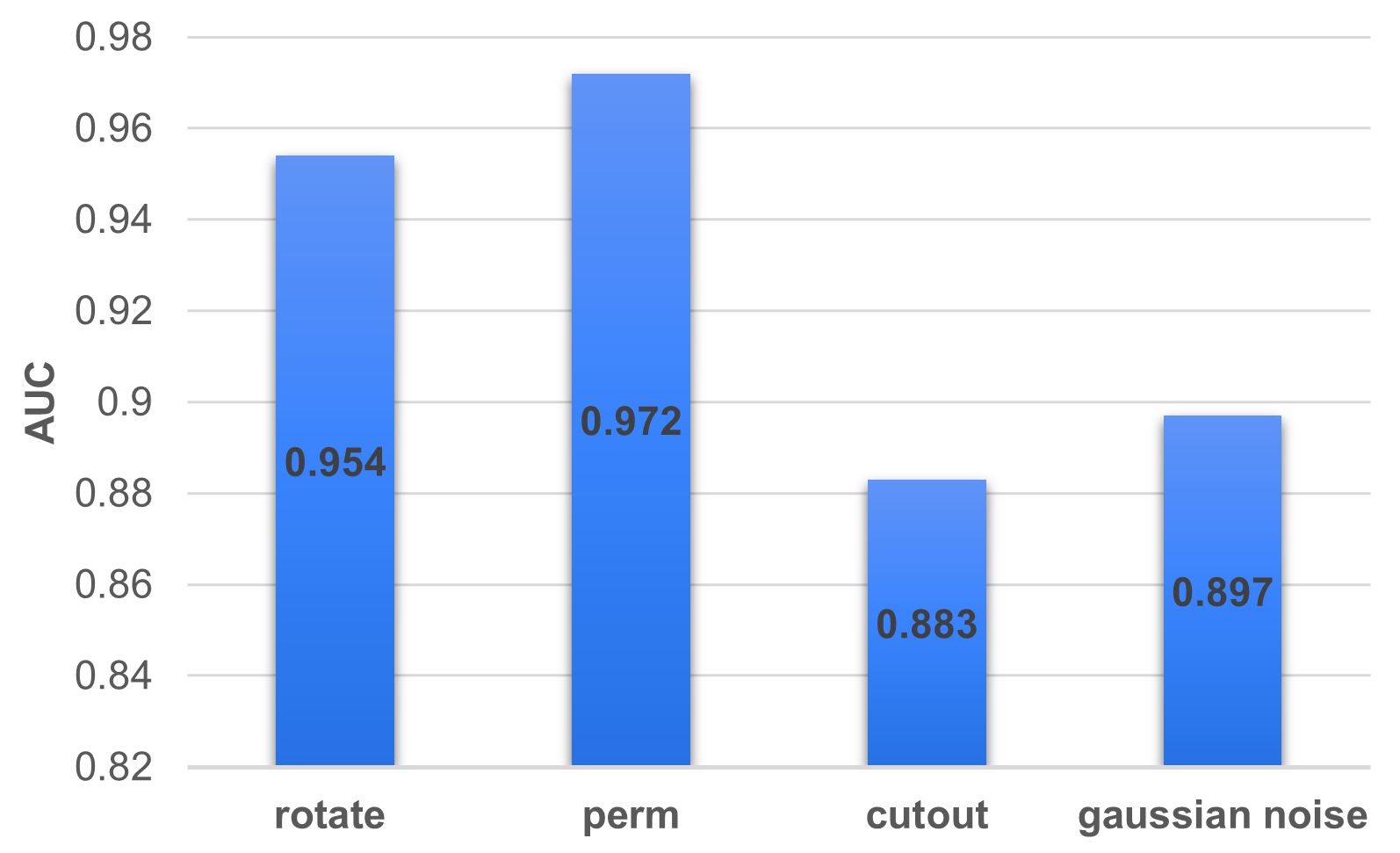}
\end{subfigure}%
 \caption{\textbf{Left}: Anomaly detection performance results based on different batch sizes of self-supervised pre-training. \textbf{Right}: Anomaly detection performance in terms of different types of strong augmentations. Both results are on Hyper-Kvasir test set using IGD as anomaly detector. 
 }
\label{fig:batch_transform_results}
\end{figure}

In Fig.~\ref{fig:batch_transform_results} (right), we explore the influence of strong augmentation strategies, represented by  rotation, permutation, cutout and Gaussian noise, on the AUC results on Hyper-Kvasir dataset, based on our self-supervised pre-training with IGD as anomaly detector. 
The experiment indicates that the use of random permutations as strong augmentations yields the best AUC results. 
We also explore the relation between batch size and AUC results in Fig.~\ref{fig:batch_transform_results} (left). The results suggest that 
small batch size (equal to 16) leads to a relatively low AUC, which increases for batch size 32, and then decreases for larger batch sizes.
Given these results, we use permutation as the strong augmentation for colonoscopy images and training batch size is set to 32.
For the LAG dataset, we omit the results, but we use rotation as the strong augmentation because it produced the largest AUC. We also used batch size of 32 for the LAG dataset.

We also present an ablation study that shows the influence of each loss term in~\eqref{eq:self_supervised_training_loss} in Tab.~\ref{tab:ablation}, again on Hyper-Kvasir dataset, based on our self-supervised pre-training with IGD. 
The vanilla contrastive learning in~\cite{simclr,moco} only achieves 91.3\% of AUC. After replacing it with our distribution contrastive loss from~\eqref{eq:contrastive_loss}, the performance increases by 2.4\% AUC. Adding distribution classification and patch position prediction losses boosts the performance by another 2.7\% and 0.8\% AUC, respectively.

\subsection{Comparison to SOTA Models}

\begin{table}[t]
\scalebox{0.85}{
\parbox{.55\linewidth}{
\centering
\begin{tabular}{cccc|c}
\toprule\hline
$\ell_{con}$\cite{simclr,moco} & $\ell_{con}$ & $\ell_{pre}$ & $\ell_{pat}$ &  AUC - Hyper-Kvasir \\ \hline \hline
\checkmark       &           &     &                    & 0.913                    \\
      & \checkmark         &     &               & 0.937                  \\
      & \checkmark   & \checkmark  &             & 0.964              \\    \hline
  & \checkmark    & \checkmark  & \checkmark       & \textbf{0.972} \\ \hline\bottomrule
\end{tabular}%
\caption{\textbf{Ablation study of the loss terms in~\eqref{eq:self_supervised_training_loss}} on Hyper-Kvasir, using IGD as anomaly detector.}
\label{tab:ablation}
}
\hfill
\parbox{.55\linewidth}{
\centering
\begin{tabular}{@{}ccc@{}}
\toprule \hline
Supervision  & Methods & Localisation - IoU   \\ \hline \hline
\multirow{4}{*}{Supervised}             & U-Net~\cite{ronneberger2015u}            & 0.746 \\
             & U-Net++~\cite{zhou2018unet++}          & 0.743 \\
             & ResUNet~\cite{diakogiannis2020resunet}          & \textbf{0.793} \\
             & SFA~\cite{fang2019selective}              & 0.611 \\ \hline
  \multirow{3}{*}{Unsupervised}            & RotNet~\cite{golan2018deep}+IGD~\cite{chen2021unsupervised}*     & 0.276  \\
             & CAVGA-$R_{u}$~\cite{venkataramanan2020attention}     & 0.349  \\
 & Ours - IGD       &  \textbf{0.372} \\ \bottomrule \hline
\end{tabular}%
\caption{\textbf{Anomaly localisation:} Mean IoU results on Hyper-Kvasir on 5 different groups of 100 images with ground truth masks. * indicates that we pretrained the geometric transformation-based anomaly detection~\cite{golan2018deep} using IGD~\cite{chen2021unsupervised} as the UAD method.}
\label{tab:localisation_auc_HK}
}

}
\end{table}

In Tab.~\ref{tab:detection_auc_HK}, we show the results of anomaly detection on Hyper-Kvasir, Liu et al.'s colonoscopy dataset and LAG datasets. 
The IGD, F-anoGAN and MS-SSIM methods improve their baselines (without our self-supervision method) from 3.3\% to  5.1\% of AUC on Hyper-Kvasir, from -0.3\% to 12.2\% on Liu et al.'s dataset, and from 0.9\% to 7.8\% on LAG. 
The IGD with our pre-trained features achieves SOTA anomaly detection AUC on all three datasets. Such results suggest that our self-supervised pre-training can effectively produce good representations for various types of anomaly detectors and datasets.  
OCGAN~\cite{perera2019ocgan} constrained the latent space based on two discriminators to force the latent representations of normal data to fall at a bounded area. CAVGA-$R_{u}$~\cite{venkataramanan2020attention} is a recently proposed approach for anomaly detection and localisation that uses an attention expansion loss to encourage the model to focus on normal object regions in the images. 
These two methods achieve 81.3\% and 92.8\% AUC on Hyper-Kvasir, respectively, which are well behind our self-supervised pre-training with IGD of 97.2\% AUC. 

We also investigate the anomaly localisation performance on Hyper-Kvasir in Tab.~\ref{tab:localisation_auc_HK}.
Compared to the SOTA UAD localisation method, CAVGA-$R_{u}$~\cite{venkataramanan2020attention}, our approach with IGD is more than 3\% better in terms of IoU.
We also compare our results to \textbf{fully supervised methods}~\cite{ronneberger2015u,zhou2018unet++,diakogiannis2020resunet,fang2019selective} to assess how much performance is lost by suppressing supervision from abnormal data.
The fully supervised baselines~\cite{ronneberger2015u,zhou2018unet++,diakogiannis2020resunet,fang2019selective} use 80\% of the annotated 1,000 colonoscopy images containing polyps during training, and 10\% for validation and 10\% for testing. 
We validate our approach using the same number of testing samples, but without using abnormal samples for training. 
The localisation results are post processed by the Connected Component Analysis (CCA)~\cite{chai1999significance}.
Notice on Tab.~\ref{tab:localisation_auc_HK} that we lose between $0.3$ and $0.4$ IoU for not using  abnormal samples for training.

\begin{table}[t]
\centering
\scalebox{0.8}{
\begin{tabular}{@{}c|c|c|c@{}}
\toprule \hline
Methods         & Hyper - AUC & Liu et al. - AUC & LAG - AUC \\ \hline\hline
DAE~\cite{masci2011stacked}             & 0.705  & 0.629  *    & -   \\
OCGAN~\cite{perera2019ocgan}           & 0.813    & 0.592 *    &  - \\
F-anoGAN~\cite{F-anoGAN}        & 0.907      & 0.691 *  & 0.778  \\
ADGAN~\cite{liuphotoshopping}           & 0.913     & 0.730 * &   -    \\
CAVGA-$R_{u}$~\cite{venkataramanan2020attention}     & 0.928   & -    & -       \\
MS-SSIM~\cite{chen2021unsupervised}        & 0.917    & 0.799 & 0.823        \\
IGD~\cite{chen2021unsupervised}            & 0.939 & 0.787  & 0.796        \\ 
RotNet~\cite{golan2018deep}+IGD~\cite{chen2021unsupervised}     & 0.905  & -  & - \\  
\hline
Ours - MS-SSIM  & 0.945   & 0.796   & 0.839      \\
Ours - F-anoGAN & 0.958     & 0.813  &   0.787   \\
Ours - IGD      & \textbf{0.972}   & \textbf{0.837}     & \textbf{0.874}   \\ \hline\bottomrule
\end{tabular}
}
\caption{\textbf{Anomaly detection:} AUC results on Hyper-Kvasir, Liu et al.'s colonocopy and LAG, respectively. * indicates that the model does not use Imagenet pre-training.}
\label{tab:detection_auc_HK}
\end{table}

\begin{figure}[t]
\begin{center}
\includegraphics[width=0.75 \textwidth]{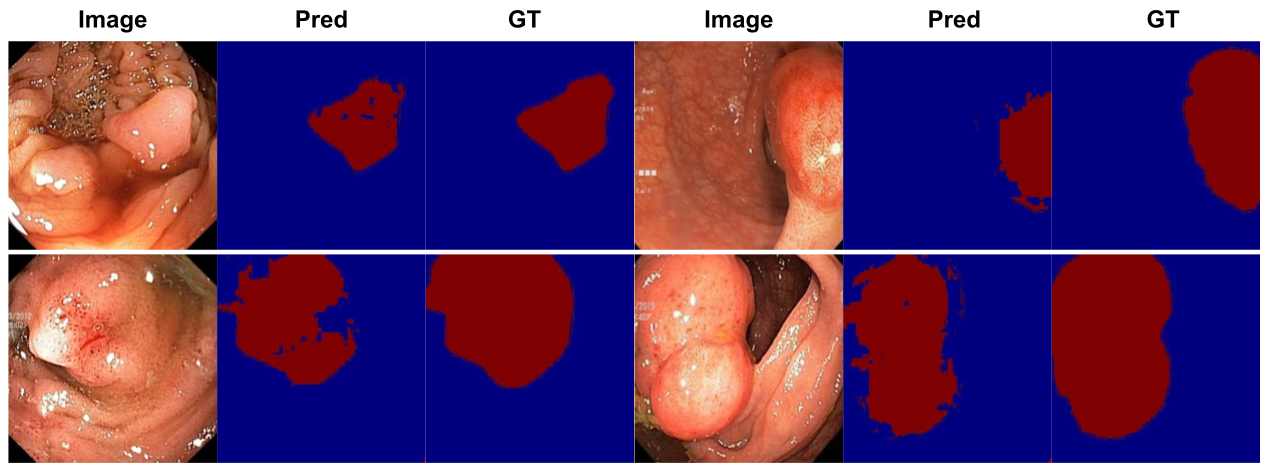}
\end{center}
\vspace{-.2in}
   \caption{Qualitative results of our localisation network based on IGD with self-supervised pre-training on the abnormal images from Hyper Kvasir~\cite{borgli2020hyperkvasir} test set.}
\label{fig:heatmap}
\end{figure}

%





We present visual anomaly localisation results of our IGD with self-supervised pre-training on the abnormal images from Hyper Kvasir~\cite{borgli2020hyperkvasir} test set in Fig.~\ref{fig:heatmap}. Notice how our model can accurately localise  polyps with various size and textures.

\section{Conclusion}

To conclude, we proposed a self-supervised pre-training for UAD named as constrained contrastive distribution learning for anomaly detection. Our approach enforces non-uniform representation distribution by constraining contrastive distribution learning with two pretext tasks. We validate our approach on three medical imaging benchmarks and achieve SOTA anomaly detection and localisation results using three UAD methods. In future work, we will investigate more choices of pretext tasks for UAD.


\bibliographystyle{splncs04}
%
\bibliography{bibli}

\end{document}